\documentclass[10pt,twocolumn,letterpaper]{article}

\usepackage[pagenumbers]{cvpr} 

\usepackage{graphicx}
\usepackage{amsmath}
\usepackage{amssymb}
\usepackage{booktabs}
\usepackage[ruled]{algorithm2e}

\usepackage{cancel} 

\usepackage{animate} 
\usepackage[accsupp]{axessibility}  

\usepackage{color}
\definecolor{citecolor}{RGB}{34,139,34}
\definecolor{grayDark}{gray}{0.95}
\definecolor{grayLight}{gray}{0.98}
\definecolor{darkgreen}{rgb}{0.8, 0.1, 0.1}

\usepackage[pagebackref=true,breaklinks=true,letterpaper=true,colorlinks,
  citecolor=citecolor,bookmarks=false]{hyperref}

\usepackage[capitalize]{cleveref}
\crefname{section}{Sec.}{Secs.}
\Crefname{section}{Section}{Sections}
\Crefname{table}{Table}{Tables}
\crefname{table}{Tab.}{Tabs.}

\def\cvprPaperID{4806} 
\def\confName{CVPR}
\def\confYear{2022}

\newcommand{\nameofmethod}{PoseTriplet}
\newcommand{\et}{\emph{et al.}}

\usepackage{color}

\begin{document}

\title{PoseTriplet: Co-evolving 3D Human Pose Estimation, Imitation, and Hallucination under Self-supervision}





\author{
{Kehong~Gong$^{1,3}$\footnotemark[1]} \qquad \quad  
{Bingbing~Li$^{2}$\footnotemark[1]}  \quad  \qquad {Jianfeng~Zhang$^1$\footnotemark[1]}  \quad  \qquad 
{Tao~Wang$^1$\thanks{Equal contribution. 
Email: {\tt \{gongkehong, zhangjianfeng, taowang\}@u.nus.edu,
l.libingbing@gmail.com}}} \quad  \qquad \\
\quad 
{Jing~Huang$^{3}$} \qquad \quad  
{Michael~Bi~Mi$^{3}$}  \quad  \qquad {Jiashi~Feng$^1$}  \quad  \qquad 
{Xinchao~Wang$^1$\footnotemark[2]\thanks{Corresponding author. Email: {\tt xinchao@nus.edu.sg}}} \quad  \qquad \\ 
\normalsize{$^{1}$National University of Singapore} \quad
\normalsize{$^{2}$Nanyang Technological University} \quad
\normalsize{$^{3}$Huawei International Pte Ltd }\\
{\tt \small{\url{https://github.com/Garfield-kh/PoseTriplet}}}.
}

\maketitle

\begin{abstract}
Existing self-supervised 3D human pose estimation schemes
have largely relied on weak supervisions like
consistency loss to guide the learning,
which, inevitably, leads to
inferior results in real-world 
scenarios with unseen poses. 
In this paper, we propose a novel self-supervised
approach that allows us to explicitly
generate 2D-3D pose pairs
for augmenting supervision,
through a self-enhancing 
dual-loop learning framework. 
This is made possible via 
introducing a
reinforcement-learning-based
imitator, which is learned jointly
with a pose estimator
alongside a pose hallucinator;
the three components
form two loops during the training process, 
complementing and strengthening one another. 
Specifically, 
the pose estimator transforms an input 2D pose sequence 
to a low-fidelity 3D output,
which is then enhanced by the imitator that
enforces physical constraints.
The refined 3D poses are subsequently 
fed to the hallucinator
for producing even more diverse data,
which are, in turn, 
strengthened by the imitator and further utilized to 
train the pose estimator. 
Such a co-evolution scheme, in practice, 
enables training a pose estimator 
on self-generated motion data without relying on any given 3D  data.
Extensive experiments 
across various  benchmarks
demonstrate that 
our approach yields encouraging 
results  significantly outperforming
the state of the art
and,
in some cases, even on par
with results of fully-supervised methods. 
{Notably, it achieves 89.1\% 3D PCK on MPI-INF-3DHP under self-supervised cross-dataset evaluation setup, improving upon the previous best self-supervised method~\cite{hu2021unsupervised, kundu2020self} by 8.6\%.}
\end{abstract}
\vspace{-2mm}
\section{Introduction}
\label{sec:intro}

\begin{figure}[!t]
\centering
\includegraphics[width=0.95\linewidth]{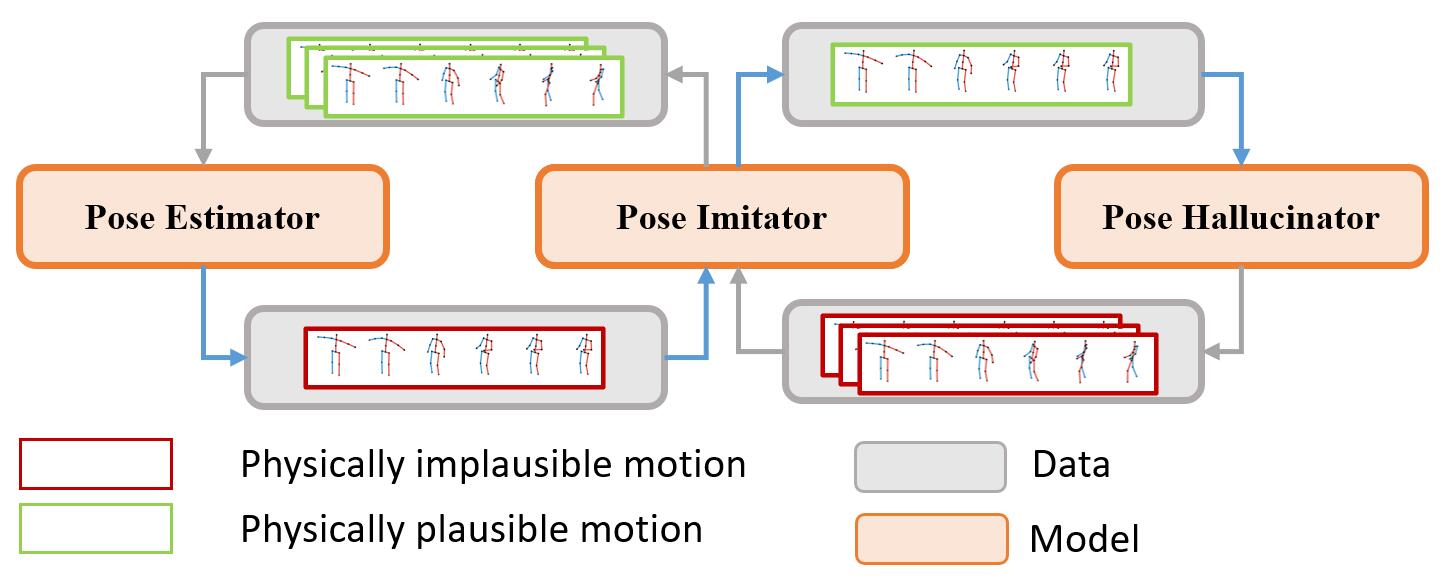}
\caption{\textbf{Overview of our \nameofmethod{} framework}. {The pose estimator, imitator, hallucinator are trained jointly in a dual-loop strategy. In the first loop, the estimator provides physically implausible motion information, which is then enhanced by the imitator via enforcing physical constraints to generate physically plausible motion. In the second loop, the hallucinator generates more diverse motion patterns given motion sequence from previous loop, and sends them to the imitator again for further refinement. This dual-loop paradigm facilitates tight co-evolution of the three components and enables iterative self-improving training of the estimator with the generated diverse and plausible motion data.}}
\vspace{-7mm}
\label{fig:summary}
\end{figure}

Video-based 3D human pose estimation 
aims to infer  3D pose sequences from videos,
and therefore plays a crucial role in many applications
such as action recognition~\cite{yan2018spatial,si2019attention}, virtual try-on~\cite{liu2021spatt}, and mixed reality~\cite{mehta2017vnect, joo2020exemplar,chengpnerf}.
Existing methods~\cite{martinez2017simple,sun2018integral,nie2019spm,kocabas2020vibe,pavllo2019videopose3d} mainly rely on 
the fully-supervised paradigms, 
in which the ground truth 3D data are given
as input. However, capturing 3D pose data 
is cost-intensive and time-consuming,
as it typically requires a multi-view 
setup or a  motion capturing system~\cite{mehta2017vnect, ionescu2014human3},
making it infeasible under in-the-wild scenarios.

To this end, two categories of methods have been introduced
to alleviate the 3D data availability issue.
The first category explores the 
semi-supervised settings, in which only a small amount
of the 3D annotations are given~\cite{zhou2017towards, li2019boosting, mitra2020multiview}.
The second category, on the other hand,
assumes no 3D data are available at all
and only 2D poses are provided.
Under this setup, 
state-of-the-art methods 
have mainly focused on imposing
weak supervision signals to guide the training,
such as aligning the projection
of an inferred 3D pose with a 2D pose~\cite{chen2019unsupervised, yu2021towards, hu2021unsupervised}.
Due to the lack of 3D data and hence
the missing of  2D-3D pairs, 
these methods are, by nature,
brittle to the challenging scenarios such as unseen poses
inherent to the in-the-wild tasks.

In this paper, we propose a novel 
self-supervised approach termed as PoseTriplet,
which allows for explicitly generating
physically- and semantically-plausible 
2D-3D pose pairs, so that full supervisions
can be imposed and further significantly strengthen the 
self-learning process.
This is made possible through introducing
a reinforcement-learning-based \emph{imitator},
which is jointly optimized with the pose \emph{estimator}
alongside a pose \emph{hallucinator}.
Specifically, the imitator takes the form
a of physics simulator with non-differentiable dynamics {to ensure physically plausibility}.
The hallucinator helps generate more diverse motion with generative motion completion.
These three key components
are integrated into a self-contained
framework and co-evolve
via a dual-loop strategy
as the training proceeds.
With only 2D pose data as input,
PoseTriplet progressively generates,
refines and hallucinates 
3D data, which in turn reinforces
all components in the loop.
Once trained, each component 
of PoseTriplet 
can be readily taken out and serves
as an off-the-shelf 
tool for its dedicated task,
such as pose estimation or imitation. 

The key motivation behind co-evolving the 
pose estimator, imitator and hallucinator,
lies in their complementary natures. In particular,
pose estimator takes 2D poses as input and generates
3D poses with reasonable semantics~(e.g., nature behaviors) but 
implausible dynamics;
such derived 3D poses are then refined through 
the physics-based imitator that enforces physical constraints.
Conversely, 
the reinforcement-learning-based imitator is {possible} to
generate unnatural behaviors~(e.g., overly energetic movements), 
which can be rectified through the pose estimator to
ensure the semantic plausibility.
Pose hallucinator, on the other hand, 
enhances the data diversity 
by producing realistic 3D pose sequences
under both the semantic and physical guidance,
which further strengthens data 
synthesizing 
and hence {improves} generalization performance. 

We show the overall workflow of PoseTriplet in Fig.~\ref{fig:summary},
which effectively aligns with aforementioned motivation. 
Unlike prior endeavors that rely on self-consistency-based
supervisions or 3D sequences as input,
PoseTriplet,
through the dual-loop scheme,
turns the input 2D poses into 
dependable 3D poses of 
realistic semantics and dynamics,
thereby lending itself to 
much stronger supervisions
and consequently the co-evolution
of the pose estimator, 
imitator and hallucinator.
Experimental results across H36M, 3DHP, and 3DPW datasets
demonstrate that, 
PoseTriplet gives rises to 
pose estimation results significantly superior
to the state-of-the-art self-supervised methods,
and sometimes even on par with 
results from fully-supervised ones. 
{Notably, it achieves 89.1\% 3D PCK on MPI-INF-3DHP under self-supervised cross-dataset evaluation setup, improving upon the previous best self-supervised method~\cite{hu2021unsupervised, kundu2020self} by 8.6\%.}

Our contribution is therefore a novel scheme
dedicated for self-supervised 3D pose estimation,
achieved by the co-evolution of a
pose estimator, imitator, and hallucinator. 
The three components complement and benefit one another, 
together leading
to a self-contained system
that enables realist 3D pose sequences
and further the 2D-3D augmented supervisions.
By taking only 2D poses as input,
PoseTriplet delivers
truly encouraging results across various benchmarks,
largely outperforming the state of the art 
and even approaching full-supervised results.

\section{Related works}

\noindent \textbf {3D pose estimation}
3D pose estimation have been wildly explored under fully supervised, semi-supervised, self-supervised.
Various approaches have been explored under fully supervised setting~\cite{martinez2017simple,mehta2017vnect,sun2018integral, nie2019spm,Yang2020Distill,pavllo2019videopose3d,kocabas2020vibe,Yang2021Pose,zhang2021bmp,wang2021mvp,WangJueICCV19}.
Through offering impressive results, those approaches highly rely on accurate motion capture data, which are hard to collect.
To address high cost of data collection, 
semi-supervised methods~\cite{zhou2017towards, li2019boosting, mitra2020multiview} are proposed to utilize the information from unlabeled data.  
Besides semi-supervised approach, augmentation based methods~\cite{Li_2020_CVPR, gong2021poseaug} are proposed to enlarge the data amount through evolution strategy~\cite{Li_2020_CVPR} or learnable approach~\cite{gong2021poseaug}.

Different from the above schemes,   self-supervised methods, with multi-view data,    explore the intrinsic supervision for model training, without requiring  ground truth 3D pose~\cite{kocabas2019epipolar, wandt2021canonpose,iqbal2020weakly }. 
For instance, Kocabas~\et~\cite{kocabas2019epipolar} utilize the epipolar geometry to generate pseudo label, ~\cite{wandt2021canonpose, iqbal2020weakly} utilize the 3D pose consistency across different views. Though being effective, those approaches require synchronized multiple cameras, which are not usual in real scenarios.
Other methods~\cite{drover2018can, chen2019unsupervised, yu2021towards, hu2021unsupervised} explore the more challenge single view setting.
For example, Drover~\et~\cite{drover2018can} utilize the prior that a random projection of a plausible 3D pose estimation will be plausible in 2D pose distribution through adversary training.
Chen~\et~\cite{chen2019unsupervised} improves this idea by adding cycle consistency. Yu~\et~\cite{yu2021towards} further introduces the scale steps for 2D poses to resolve the ambiguity issue.
Zhang~\et~\cite{zhang2020inference} applies self-supervised learning on test data to adapt model to new scenarios.

Our method belongs to the self-supervised approaches {under single view setting}.
Different from previous self-supervised approaches which implement weak supervision signal through consistency~\cite{chen2019unsupervised} or adversary~\cite{drover2018can, yu2021towards}, our method directly uses the strong supervision signal from self-generated data, results in more accurate and stable model performance.
The pseudo label strategy~\cite{li2019boosting} under semi-supervised category is close to our approach. However, our approach does not require ground truth data for model pretraining, and our method introduces physical plausibility refinement and diversity enhancement to achieve better performance, which are absent in~\cite{li2019boosting}.

\noindent \textbf {Physics-based pose estimation}
The above methods   are all Kinematics based. Though providing impressive results, they do not consider  physical constrains, thus suffering physical implausible artifacts (\eg,  foot skating and ground penetration).
To ensure physical plausibility, recent works explore physical constraints.
Rempe~\et~\cite{rempe2020contact} introduces physical law to the foot contact and human dynamic,
while its iterative optimization is high time costly (\eg, 30 minutes for 2s clip).
Later, ~\cite{shimada2020physcap, shimada2021neural, xie2021physics} propose differentiable physical constrain to reduce the time cost.
But they only consider foot contact, making them less effective in scenarios with  other important contact (\eg, lay down, sitting with chair). 

Different from optimization based approaches, physics simulation based methods use physics simulators to provide realistic physical constrains.
DeepMimic~\cite{peng2018deepmimic} tries to imitate various motion from reference mocap data in physics engine via reinforcement learning.
SFV~\cite{peng2018sfv}   proposes to refine  the low fidelity motion data from video-based pose estimation  through imitation learning.
However,  their adopted  imitation learning requires days of training for just one clip.
Later, SimPoe~\cite{yuan2021simpoe} addresses this issue by introducing RFC~\cite{yuan2020residual} to effectively reduce the time consumption by training one policy for all motion clips.
Our method is built upon SimPoe~\cite{yuan2021simpoe} for better generalization and low time cost. However, different from those methods only using physical constraints   for post processing, our method propose to involve it in the learning loop. As such,  no mocap data for pose estimation training and imitation learning is required.

\noindent \textbf {Motion synthesis}
Motion synthesis includes non-learning based and learning based approaches.
In non-learning based approaches,
the motion graph method~\cite{kovar2008motion} first builds transition edges between different motion points based on their similarity, and then generates new motion data through traversing the graphs. 
Motion matching~\cite{buttner2015motion} searches proper future frames in motion data based on motion states in real time.
In learning based approaches,
motion perdition based methods~\cite{martinez2017human, zhang2020we, li2018convolutional, gui2018adversarial, barsoum2018hp, yuan2020dlow, pavllo2018quaternet} aim to predict future poses conditioned on previous poses.  Action generation~\cite{wang2020learning, butepage2017deep, battan2021glocalnet} aims to generate pose sequence conditioned on action labels.
Motion completion~\cite{hernandez2019human, kaufmann2020convolutional, holden2016deep, duan2021ssmc, harvey2018recurrent, harvey2020robust}  generates realistic transitions between key frames, which most relevant to our work in their aims.  The pose hallucination in our framework also aims  to generate novel motion sequence, where motion graph and motion match methods are not applicable due to tight restriction in their generated data. 
We therefore choose motion completion considering it can  generate longer sequence with continuously input key frames.
\section{Methodology}
Given a 2D pose sequence $\boldsymbol{x}_{1:T} = (\boldsymbol{x}_1,...,\boldsymbol{x}_T)$ of length $T$, where $\boldsymbol{x}_t \in \mathbb{R}^{J\times 2}$ is the 2D spatial coordinate of $J$ body joints at time $t$, our goal is to estimate the 3D pose sequence $\boldsymbol{X}_{1:T} = (\boldsymbol{X}_1,...,\boldsymbol{X}_T)$, where $\boldsymbol{X}_t \in \mathbb{R}^{J\times 3}$ is the corresponding 3D joint position under the camera coordinate system.
Conventionally, a pose estimator $\mathcal{P}: \boldsymbol{x}_{1:T} \mapsto \boldsymbol{X}_{1:T}$ with parameter $\theta$ is trained with a large set of paired 2D and 3D pose data $\{\boldsymbol{{x}}_{1:T}, \boldsymbol{{X}}_{1:T}\}$ through fully-supervised learning approaches~\cite{martinez2017simple,mehta2017vnect,sun2018integral,kocabas2020vibe, pavllo2019videopose3d}:
\begin{equation}
\min_{\theta} \mathcal{L}_{\mathcal{P}}(\mathcal{P}_{\theta}(\boldsymbol{\boldsymbol{{x}}_{1:T}}),\boldsymbol{\boldsymbol{{X}}_{1:T}}).
\end{equation}
Here $\mathcal{L}_{\mathcal{P}}$ denotes the loss function which is typically
defined as mean square errors (MSE) between predicted and ground truth 3D poses sequences.
However, ground truth 3D pose data is expensive to capture, which limits the applicability of these approaches. 
To avoid using 3D data, previous self-supervised approaches typically apply weak 2D re-projection loss~\cite{drover2018can, chen2019unsupervised, yu2021towards, hu2021unsupervised} to learn the estimator:
\begin{equation}
\min_{\theta} \mathcal{L}_{\mathcal{P}}(\Pi(\mathcal{P}_{\theta}(\boldsymbol{\boldsymbol{{x}}_{1:T}})),\boldsymbol{\boldsymbol{{x}}_{1:T}}),
\end{equation}
where $\Pi$ is the perspective projection function.
The re-projection loss only provides weak supervision which tends to induce unstable or unnatural estimations. In this work, we aim to design a self-supervised learning framework of which the core is an iterative self-improving paradigm. Specifically, we propose to enhance the current estimation with some specifically designed transformation $\mathcal{T}$ (\eg, to produce more smooth and diverse motion):
\begin{equation}
\begin{aligned} \boldsymbol{{X}}'_{1:T}=\mathcal{T}(\mathcal{P}_{{\theta}_{n}}(\boldsymbol{\boldsymbol{{x}}_{1:T}}))
\end{aligned}
\end{equation}
The enhanced estimates are then projected to 2D pose to obtain paired training data $\{\boldsymbol{{x}}'_{1:T}, \boldsymbol{{X}}'_{1:T}\}$, which are used to improve the pose estimator:
\begin{equation}
{\theta}_{n+1} \leftarrow \min_{\theta} \mathcal{L}_{\mathcal{P}}(\mathcal{P}_{\theta}(\boldsymbol{\boldsymbol{{x}}'_{1:T}}),\boldsymbol{\boldsymbol{{X}}'_{1:T}})
\end{equation}
Here $\theta_{n}$ and $\theta_{n+1}$ denote the parameters of the current estimator and improved estimator.
The improved estimator can then be utilized to start a new iteration of data enhancement and training.
Build on this \textit{self-improving} paradigm, we can train a superior pose estimator starting from only a set of 2D pose sequences $\{\boldsymbol{{x}}_{1:T}\}$.


\begin{figure*}[!t]
\centering
\includegraphics[width=0.85\linewidth]{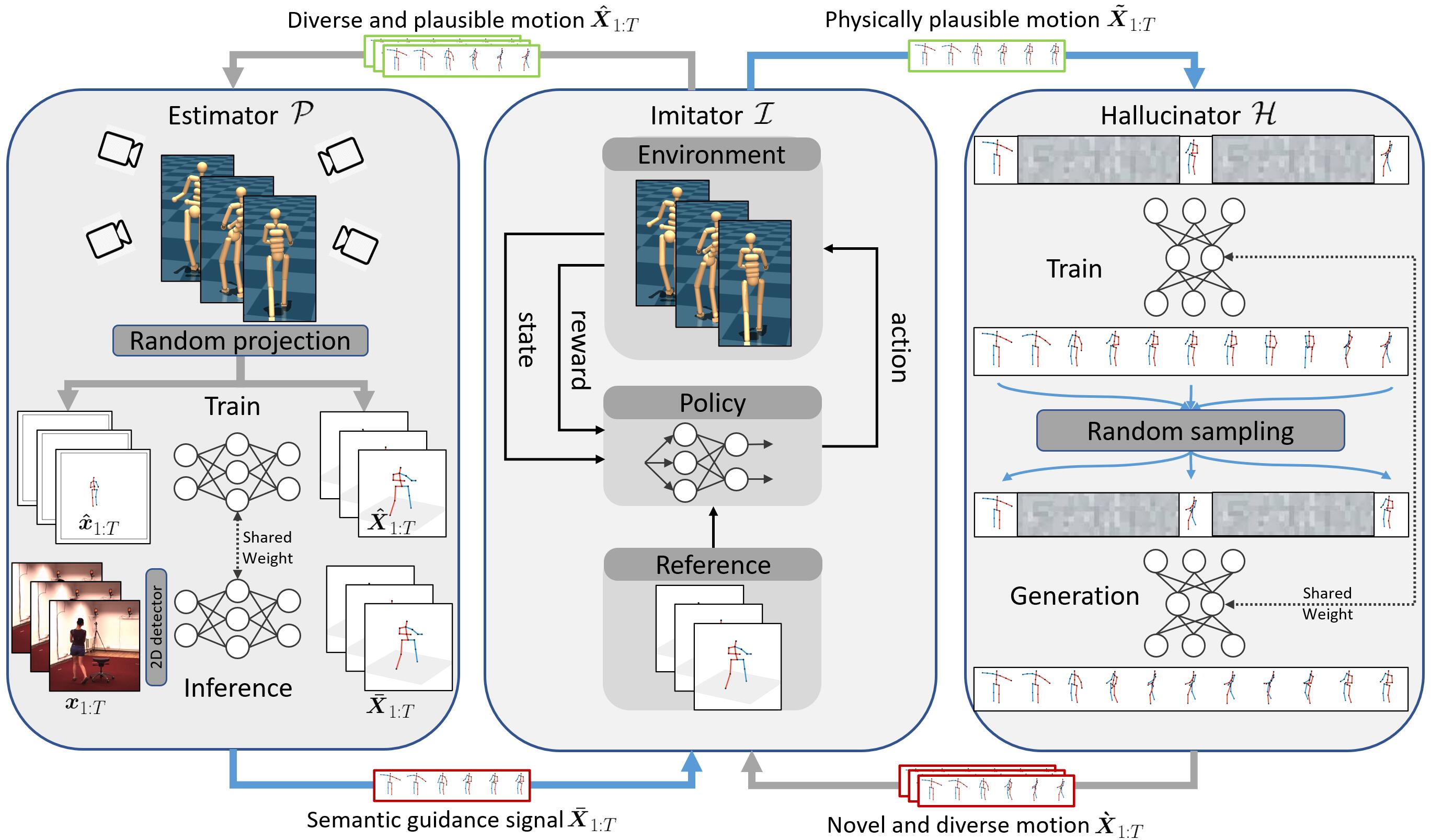}
\caption{\textbf{Detail of our \nameofmethod{} framework}. 
Given available 2D pose sequence $\boldsymbol{{x}}_{1:T}$, the pose estimator $\mathcal{P}$ transforms it to low-fidelity 3D pose sequence $\boldsymbol{\bar{X}}_{1:T}$.
$\boldsymbol{\bar{X}}_{1:T}$ is then served as semantic guidance signal (\ie, reference motion) for imitator $\mathcal{I}$ to obtain physically plausible motion $\boldsymbol{\tilde{X}}_{1:T}$.
The hallucinator $\mathcal{H}$ then generates novel and diverse motion $\boldsymbol{\grave{X}}_{1:T}$ from  $\boldsymbol{\tilde{X}}_{1:T}$, which is then refined by the imitator $\mathcal{I}$ to obtain the final enhanced diverse and plausible motion $\boldsymbol{\hat{X}}_{1:T}$.
$\boldsymbol{\hat{X}}_{1:T}$ is then projected to 2D-3D pairs to train the estimator.
The improved estimator takes the available 2D pose sequence $\boldsymbol{{x}}_{1:T}$ and start another round of dual-loop optimization.
}
\vspace{-3mm}
\label{fig:overview}
\end{figure*}

\subsection{\nameofmethod{}}

To construct an effective self-improving framework, we identify two challenging aspects for enhancing the 3D motion sequence: 1) the pose estimation from the estimator may not be physically plausible due to ignorance of force, mass and contact modeling; 2) existing 2D motion  may be limited in diversity and thus the learned model cannot generalize well. 
To address these challenges, we introduce a pose imitator based on the reinforcement learning aided human motion modeling and a pose hallucinator based on generative motion interpolation accordingly to refine and diversify the 3D motion. The former helps correct the physical artifacts while the latter generates novel pose sequences based on the existing estimates. 
We find these two aspects in motion are complementary and thus combine them together.
The resulting pipeline helps obtain 3D motion data $\{\boldsymbol{{x}}'_{1:T}, \boldsymbol{{X}}'_{1:T}\}$ with significantly improved physical plausibility and motion diversity.
Nevertheless, we find naive two-step combination of the two approaches generate  inferior-quality 3D pose sequence. 
The reason is that performing motion diversification first could be ineffective due to implausible estimate while conducting motion diversification later could introduce physical artifacts.
Therefore, we further introduce a \textit{dual-loop} scheme and unify the two components with pose estimator into a novel self-supervised framework named \nameofmethod{}.

\noindent \textbf{Dual-loop architecture}
Concretely, as shown in Fig.~\ref{fig:overview}, our \nameofmethod{} introduces a dual-loop architecture to integrate the three modules: a \textit{pose estimator} $\mathcal{P}$, a \textit{pose imitator} $\mathcal{I}$, and a \textit{pose hallucinator} $\mathcal{H}$.
Given the set of available 2D pose sequence $\boldsymbol{{x}}_{1:T}$,
the pose estimator first transforms them to low-fidelity 3D pose sequence:
\begin{equation}
\vspace{-1mm}
\begin{aligned} \boldsymbol{\bar{X}}_{1:T}=\mathcal{P}(\boldsymbol{\boldsymbol{{x}}_{1:T}})
\end{aligned}
\end{equation}
$\{\boldsymbol{\bar{X}}_{1:T}\}$ is converted to low-fidelity reference motions and served as semantic guidance signal to the pose imitator, which imposes the physical human motion dynamic modeling and obtains physically plausible motion sequence:
\begin{equation}
\vspace{-1mm}
\begin{aligned} \boldsymbol{\tilde{X}}_{1:T}=\mathcal{I}(\boldsymbol{\boldsymbol{\bar{X}}_{1:T}})
\end{aligned}
\end{equation}
By learning a generative motion completion model, the pose hallucinator then generates novel and diverse motion sequences $\{\boldsymbol{\grave{X}}_{1:T}\}$ based on the improved plausible motion from the imitator:
\begin{equation}
\vspace{-1mm}
\begin{aligned} \boldsymbol{\grave{X}}_{1:T}=\mathcal{H}(\boldsymbol{\boldsymbol{\tilde{X}}_{1:T}})
\end{aligned}
\end{equation}
Afterwords, instead of closing the loop by treating $\{\boldsymbol{\grave{X}}_{1:T}\}$ as augmented data to the estimator, we introduce another loop.
We feed $\{\boldsymbol{\grave{X}}_{1:T}\}$ back into the imitator to correct the induced physical artifacts and obtain the final expected plausible and diverse motion sequences:
\begin{equation}
\vspace{-1mm}
\begin{aligned} \boldsymbol{\hat{X}}_{1:T}=\mathcal{I}(\boldsymbol{\boldsymbol{\grave{X}}_{1:T}})
\end{aligned}
\end{equation}
$\{\boldsymbol{\hat{X}}_{1:T}\}$ is then projected to 2D to obtain paired data $\{\boldsymbol{\hat{x}}_{1:T}, \boldsymbol{\hat{X}}_{1:T}\}$ for training the pose estimator.

By jointly optimizing this dual-loop architecture, the three components form a tight co-evolving paradigm: 
1) the estimator benefits from the diverse and plausible augmented data to learn more accurate estimation.
2) the imitator learns more robust and physically natural motion based on the improved estimation and diverse data generated from the hallucinator.
3) the hallucinator generates diverse pose sequence of higher quality based on the improved data from the imitator.

\noindent \textbf{Loop starting}
Another challenging aspect of this self-improving learning paradigm is the loop starting. Without access to 3D motion data, the whole framework cannot start learning. Recall our pose imitator employs physics-based human motion model, we thus develop a zero-data generating strategy that produces initial 3D pose sequence for starting the dual-loop learning.
Specifically, we generate root trajectory signal in horizontal plane with random direction and proper velocity.
This trajectory is then used for guidance signal for RL agent.
By control the agent to follow the generated trajectory, we can generate motion sequences that are physically plausible.
These motion sequences is then projected to obtain 2D-3D pose pairs and used to train a initial pose estimator. 
In this way, the whole dual-loop learning can be started.


\subsection{Module detail}


\subsubsection{Pose estimator}
The pose estimator estimates the 3D pose sequence $\boldsymbol{X}_{1:T}$ from the input sequence $\boldsymbol{x}_{1:T}$. 
Specifically, we adopt a similar estimator architecture as VideoPose~\cite{pavllo2019videopose3d}, which predicts both root trajectory and root-relative joint locations. The trajectory can be used as additional movement signal to pose imitator.
Meanwhile, the noise in root movement can be corrected by the pose imitator and in turn help the pose estimator.
We use Mean Square Error (MSE) loss for the root-related pose estimation and Weighted L1 loss for the trajectory estimation following~\cite{pavllo2019videopose3d}.

\noindent \textbf{Projection for training estimator}
Given the generated motion sequence data $\{\boldsymbol{\hat{X}}_{1:T}\}$, we project them to 2D to obtain paired training data.
We consider two strategies for the projection: 
1) Heuristic random projection.
We set the virtual camera with certain elevation, azimuth range, height and distance range to match the {indoor capture environment}. This is similar to the projection strategy for 3D pose data synthesis as Chen~\et\cite{chen2016synthesizing};
2) Generative adversarial learning based projection~\cite{gong2021poseaug}. A generator is used to regress the camera orientation and position for each motion sequence. The regression is learned through a discriminator by distinguishing the real and the projected 2D pose sequences with the generated camera parameters.
In this way, reasonable camera viewpoint distribution can be extracted from real 2D pose data, improving the plausibility of generated 2D-3D paired data.
The two strategies are combined in our framework to ensure the diversity of camera viewpoints.

\subsubsection{Pose imitator}
The 3D pose sequences $\{\boldsymbol{\bar{X}}_{1:T}\}$ predicted from pose estimator  $\mathcal{P}$, due to lack of physical constrain, would suffer unnatural artifacts such as foot skating, floating, floor penetration. 
Those artifacts prevent it from being used as training data directly for estimator $\mathcal{P}$ or hallucinator $\mathcal{H}$.
To address the issue, motivated by \cite{peng2018deepmimic,peng2018sfv, yuan2021simpoe}, we introduce a reinforcement learning based pose imitator  $\mathcal{I}$ to imitate the low fidelity 3D pose sequence $\{\boldsymbol{\bar{X}}_{1:T}\}$ from pose estimator to generate  more physically plausible motion sequence $\{\boldsymbol{\tilde{X}}_{1:T}\}$.

\noindent \textbf{Background}
The imitation process can be seen as a Markov decision process.
Given a reference motion and current state $\boldsymbol{s}_t \in \mathcal{S}$, the agent interacts with the simulation environment with action $\boldsymbol{a}_t \in \mathcal{A}$ and receive reward $\boldsymbol{r}_{t}$.
the action is determined by a policy $\pi(\boldsymbol{a}_t|\boldsymbol{s}_t)$ conditioned on state $\boldsymbol{s}_t \in \mathcal{S}$;
the reward is determined based on how similar the agent behaves like the reference motion.
When an action is taken, the current state $\boldsymbol{s}_t$ changes to next state $\boldsymbol{s}_{t+1}$ through transition function $\mathcal{T}(\boldsymbol{s}_{t+1}|\boldsymbol{s}_{t}, \boldsymbol{a}_{t})$.
The goal is to learn a policy that maximizes the average cumulative rewards $\sum_{i=1}^{\infty} \boldsymbol{\gamma}^{i} \boldsymbol{r}_{t}$ (\ie, performing similar behavior in physics simulator as reference motion), where $\boldsymbol{\gamma}$ is the discounting factor. The state, action and rewards are detailed below.

\noindent \textbf{State} includes current pose $\boldsymbol{q}_{t}$, current velocity $\dot{\boldsymbol{q}_{t}}$, and target pose $\widetilde{\boldsymbol{q}}_{t+1}$ from reference motion. 
To deal with the noisy reference motion from the pose estimator, we introduce an extra encoded feature $\phi$ by concatenating and fusing the past and future motion information.
In this way, the control policy is aware of past and future reference motion, and is thus more robust to the noise.

\noindent \textbf{Action} involves two kinds of forces: internal force and external force. 
The internal force is applied by actuator on the non-root joints (\eg, elbow, knee). Following previous work~\cite{peng2017learning}, we use PD (proportional–derivative) control for internal force control.
The external force $\boldsymbol{\eta}_{t}$ is a virtual force applied on root joint(\ie, hip)~\cite{yuan2020residual} for extra interaction (\eg, sitting on the chair) and is regressed by the policy network.

\noindent \textbf{Rewards} measure the motion differences between the agent and reference motion. 
These differences capture pose related (pose, velocity), root related (root height, root velocity) and body end factors (position, velocity).
Besides, a regulation loss on virtual force is applied to avoid unnecessary external force following \cite{yuan2020residual}.
As we find that the agent is hard to move with the above setting due to the noisy reference motion, we further introduce a feet relative position into the motion characteristics
to enhance the feet motion.

\subsubsection{Pose hallucinator}
The pose hallucinator aims to generate novel and diverse motion sequence based on the refined data from pose imitator. 
In this work, we choose motion interpolation technique to generate novel pose motions. Specifically, we sample key-frames from the refined pose sequence, and interpolate the missing frames via neural networks to generate new motion data.
In details, the pose hallucinator is constructed by a recurrent neural network (RNN) structure. The inputs are the sampled temporal key-frames (we sample key-frames with a certain frame interval).
Conditioned on these sampled key-frames, the model predicts the intermediate frames in sequential manner.
A reconstruction loss and an adversary loss is used to train this model.
The reconstruction loss measures the $L_2$ distance between the ground truth and predicted poses. The adversary loss provides temporal supervision to avoid RNN collapse (\ie, predicting average motion).
In the inference stage, we randomly select frames from different motion clips and generate novel motion sequences based on these sampled key-frames.

\section{Experiments}

We study three questions in experiments.
1)  Is \nameofmethod{} able to improve performance of video pose estimator for both intra- and  cross-dataset scenarios?
2) How does the performance improves with the round of co-evolving process?
3) How does the amount of training data affects model performance?
We conduct experiments with H36M (source dataset) and 3DHP/3DPW (for cross-dataset evaluation). 
Throughout the experiments, we adopt VideoPose~\cite{pavllo2019videopose3d} (T=27) as our pose estimator. We report results from estimator for comparison. Please refer to supplementary for more implementation details.

\subsection{Dataset}
\noindent \textbf{H36M}~\cite{ionescu2014human3} 
is the most popular 3D pose benchmark captured by marker-based motion capture system.
It contains 3.6 million video frames for 11 subjects and 15 scenarios.
Following previous works~\cite{chen2019unsupervised, yu2021towards}, we use the 2D poses of subject S1, S5, S6, S7, S8 as our training set and evaluate the performance on S9 and S11.
The two standard metrics Mean Per Joint Position Error (MPJPE) in millimeters and Procrustes Aligned Mean Per Joint Position Error (PA-MPJPE) are used for evaluation.

\noindent \textbf{3DHP}~\cite{mehta2017vnect} is a large 3D pose dataset. It contains both indoor and outdoor scenarios.
Following previous works~\cite{kolotouros2019spin, chen2019unsupervised}, we report the metrics of MPJPE, Percentage of Correct Keypoints (PCK) and Area Under the Curve (AUC) after scale and rigid alignment for evaluation.
We only use its test set to evaluate the model's generalization performance.

\noindent \textbf{3DPW}~\cite{vonMarcard2018} is a more challenging in-the-wild dataset.
It contains more complicated activities and scenarios.
Same as 3DHP, we only use its test set to evaluate model's generalization performance.
Follow previous work~\cite{kolotouros2019spin}, we report MPJPE and PA-MPJPE for 3DPW.

\subsection{Quantitative results}


\noindent \textbf{Results on H36M} 
We compare our \nameofmethod{} with other state-of-the-art self-supervised methods~\cite{rhodin2018unsupervised, chen2019unsupervised, kundu2020self, yu2021towards, hu2021unsupervised} under GT (ground truth 2D poses) and Det (detected 2D poses) settings as shown in Table~\ref{tab:h36m}.
Among which, 
\cite{rhodin2018unsupervised,chen2019unsupervised,kundu2020self} implement weak supervision (\ie, consistency supervision),
\cite{yu2021towards,hu2021unsupervised} utilize  temporal information through adversary learning~\cite{yu2021towards} and smoothness constrains~\cite{hu2021unsupervised}.
Our method outperforms the best of them by a large margin in MPJPE for both GT (85.3 \vs 68.2) and Det (82.1 \vs 78.0) settings.
The result verifies that our method with co-evolving strategy and augmented supervision performs better compared with previous approaches.
{Moreover, our method also outperforms some weakly-supervised approaches~\cite{wu2016single, tung_aign, li2019boosting, iqbal2020weakly} which involve ground truth data during training. 
Especially, comparing with Li~\et~\cite{li2019boosting} which implements low rank representation and temporal smoothing for pseudo 3D label generation,
our approach, utilizing the advantage of physics simulator, provides better refinement and outperforms \cite{li2019boosting} by a large margin in MPJPE (88.8 Vs 78.8) even it use ground truth data (\ie, subject 1). 
This verifies the effectiveness of our co-evolving strategy on reducing reliance on 3D data.}

\begin{table}[h]
	\small
	\centering
	\setlength{\tabcolsep}{1mm}
	\begin{tabular}{l|l|cc|cc}
		\specialrule{1pt}{1pt}{1pt}
		Mode & Method &  \multicolumn{2}{c|}{GT} & \multicolumn{2}{c}{Det}  \\
		&  &  P1~($\downarrow$) & P2~($\downarrow$) &  P1~($\downarrow$) & P2~($\downarrow$)  \\
		\hline
		Full  & Martinez~\et~\cite{martinez2017simple} & 45.5 &  37.1 & 62.9 & 47.7  \\ 
		Full  & Pavllo~\et~\cite{pavllo2019videopose3d} & 37.2 & 27.2 & 46.8 & 36.5 \\ 
		\hline
		Weak & 3DInterpreter~\cite{wu2016single} & - & 88.6 & - & 98.4 \\  
		Weak & AIGN~\cite{tung_aign} & - & 79.0 & - & 97.4 \\ 
		Weak & Drover~\et~\cite{drover2018can} & - & 38.2 & - & 64.6 \\  
		Weak & Li~\et~\cite{li2019boosting} & - & - & 88.8 & 66.5 \\  
		Weak & Umar~\et~\cite{iqbal2020weakly} & - & - & - & 55.9 \\ 
		\hline
		Self & Rhodin~\et~\cite{rhodin2018unsupervised} & - & - & 131.7 & 98.2 \\ 
		Self & Chen~\et~\cite{chen2019unsupervised} & - & 51.0 & - & 68.0 \\ 
		Self & Kundu~\et~\cite{kundu2020self} & - & - & - & 62.4 \\ 
		Self & Kundu~\et~\cite{kundu2020kinematic} & - & - & - & 63.8 \\ 
		Self & Yu~\et~\cite{yu2021towards} & 85.3 & \textbf{42.0} & 92.4 & 52.3 \\ 
		Self & Hu~\et~\cite{hu2021unsupervised} & - & - & 82.1 & - \\ 
		Self & Wandt~\et~\cite{wandt2021canonpose}$^*$ & - & - & 81.9 & 53.0 \\ 
		Self & Ours   & \textbf{68.2}  & 45.1 & \textbf{78.0} & \textbf{51.8} \\	
		\specialrule{1pt}{1pt}{2pt}	
	\end{tabular}
	\vspace{-1mm}
	\caption{\textbf{Results on H36M} in terms of MPJPE~(P1) and PA-MPJPE~(P2). {$^*$ uses multi-view setting.} Best results are shown in \textbf{bold} under self supervised setting.}
	\label{tab:h36m}

\end{table}

\noindent \textbf{Results on 3DHP} 
We then evaluate the generalization performance of our method on cross dataset 3DHP. 
We compare our \nameofmethod{} with state-of-the-art methods, including fully supervised, weakly supervised, and self supervised approaches~\cite{mehta2017vnect, sun2019human, kolotouros2019spin, chen2019unsupervised, kundu2020kinematic, kundu2020self, yu2021towards}.
As shown in Table~\ref{tab:3dhp}, under cross-data evaluation, our method overruns previous self-supervised methods~\cite{chen2019unsupervised, kundu2020self, yu2021towards} significantly in PCK (82.2 \vs 89.1) and MPJPE (103.8 \vs 79.5).
The result indicates that the diverse and plausible motion generated by our \nameofmethod{} improves generalization.
Exception is that Kundu~\et~\cite{kundu2020self}, which uses extra data and unpaired 3D poses for model training and thus achieves slightly better performance in AUC (56.3 \vs 53.1).
Our method also outperforms self-supervised methods~\cite{chen2019unsupervised, kundu2020kinematic, kundu2020self, yu2021towards} trained on 3DHP dataset directly.
In addition, our method achieves better performance than weakly supervised approaches~\cite{sun2019human, kolotouros2019spin} in all metrics, even though they use unpaired images and 3D poses for supervision during the training process.
Notably, our method even achieves comparable performance with fully supervised approaches~\cite{mehta2017vnect, sun2019human, kolotouros2019spin}).
{In summary, the cross-dataset performance of our self-supervised framework \nameofmethod{} is comparable with the intra-dataset result from fully/semi-supervision. This indicates a good generalization performance of our \nameofmethod{}.}

\begin{table}[h]
	\small
	\centering
	\setlength{\tabcolsep}{1mm}
	\begin{tabular}{l|c|c|c|c|c}
		\specialrule{1pt}{1pt}{1pt}
		Mode & Method  & CE &  PCK~($\uparrow$) & AUC~($\uparrow$) & MPJPE~($\downarrow$)  \\
		\hline
		Full & VNect~\cite{mehta2017vnect} & & 83.9 & 47.3 & 98.0 \\ 
		Full & HMR~\cite{sun2019human} & & 86.3 & 47.8 & 89.8 \\ 
		Full & SPIN~\cite{kolotouros2019spin} & & 92.5 & 55.6 & 67.5 \\ 
	    \hline
	    Weak & HMR~\cite{sun2019human} & & 77.1 & 40.7 & 113.2 \\ 
	    Weak & SPIN~\cite{kolotouros2019spin} & & 87.0 & 48.5 & 80.4 \\ 
		\hline
		Self & Chen~\et~\cite{chen2019unsupervised} &  & 71.1 & 36.3 & - \\ 
		Self & Kundu~\et~\cite{kundu2020kinematic} &  & 80.2 & 44.8 & 97.1\\ 
		Self & Kundu~\et~\cite{kundu2020self} &  & 84.6 & 60.8 & 93.9 \\ 
		Self & Yu~\et~\cite{yu2021towards} & & 86.2 & 51.7 & - \\ 
		\hline
		Self & Chen~\et~\cite{chen2019unsupervised} & \checkmark & 64.3 & 31.6 & - \\ 
		Self & Kundu~\et~\cite{kundu2020self}$^*$ & \checkmark & 82.1 & \textbf{56.3} & 103.8 \\ 
		Self & Yu~\et~\cite{yu2021towards} & \checkmark  & 82.2 & 46.6 & - \\ 
		Self & Ours & \checkmark & \textbf{89.1}  & 53.1 & \textbf{79.5}  \\
		\specialrule{1pt}{1pt}{2pt}	
	\end{tabular}
	\vspace{-1mm}
	\caption{\textbf{Results on 3DHP} in terms of PCK, AUC, and MPJPE. CE denotes cross-data evaluation. {$^*$ uses extra unpaired 2D/3D dataset for training.} Best results are shown in \textbf{bold}.}
	\label{tab:3dhp}

\end{table} 

\noindent \textbf{Results on 3DPW} 
We further evaluate the generalization performance of our method on in-the-wild 3DPW dataset. 
Note that there is few works evaluated on 3DPW under the self-supervised cross dataset setting. 
Therefore, we compare to the supervised approaches~\cite{kolotouros2019spin, pavllo2019videopose3d, sun2019human, wang2020predicting} directly.
From Table~\ref{tab:3dpw}, we can observe our method achieves comparable results with the fully supervised baseline without relying on any 3D data.
{This demonstrates that our method performs well on complicated and challenging in-the-wild scenarios.} 

\begin{table}[h]
	\small
	\centering
	\setlength{\tabcolsep}{1mm}
	\begin{tabular}{l|c|c|c|c}
		\specialrule{1pt}{1pt}{1pt}
		Mode & Method  & CE & MPJPE~($\downarrow$) & P-MPJPE~($\downarrow$)  \\
		\hline
		Full & Wang~\et\cite{wang2020predicting} & \checkmark & 124.2 & - \\
		Full & DSD-SATN~\cite{sun2019human} & \checkmark & - & 69.5 \\ 
		Full & CRMH~\cite{jiang2020coherent} & \checkmark & 105.3 & 62.3 \\
		Full & BMP~\cite{zhang2021bmp} & \checkmark & 104.1 & 63.8 \\ 
		Full & VideoPose~\cite{pavllo2019videopose3d} & \checkmark & 101.8 & 63.0 \\ 
		\hline
		Self & Ours & \checkmark & 115.0  & 69.5   \\
		\specialrule{1pt}{1pt}{2pt}	
	\end{tabular}
	\vspace{-1mm}
	\caption{\textbf{Results on 3DPW} in terms of MPJPE and PA-MPJPE. CE denotes cross-data evaluation.}
	\label{tab:3dpw}

\end{table}

\subsection{Qualitative results}
While previous self-supervised methods rely on weak supervision signal (\eg, consistency loss), our method trains the pose estimator with augmenting supervision from the self-generated data, resulting in more stable, plausible, and accurate estimation\footnote{Fig.3-8 are video figure that are best viewed in Adobe Reader (click and play), and videos are in supplementary materials.}.
As shown in Fig.~\ref{fig:compare-Hu}, although Hu~\et~\cite{hu2021unsupervised} implements temporal smoothness prior during the training process, jittering effect is still obvious. While our result, learned from co-evolving approach, is much smoother.
Yu~\et~\cite{yu2021towards} introduce a scale estimation strategy for 2D pose to reduce the scale ambiguity. 
Through the weak supervision from the bone length consistency and scale distribution, 
his result still contains scale ambiguity (\ie, the body size varies) as shown in Fig.~\ref{fig:compare-yu}. 
Ours result maintains stable and accurate in term of body size compared with it.
We further demonstrate the result from 3DHP (Fig.~\ref{fig:vis-3dhp}) and 3DPW (Fig.~\ref{fig:vis-3dpw}). These results demonstrate that our method perform well on unseen poses for in-the-wild scenarios.
More in-the-wild examples can be viewed in the supplementary material in video format. 


\begin{figure}[!t]
\centering
\animategraphics[scale=0.37]{10}{gif/iccv2021hu/iccv2021dance_}{12}{32}
\caption{\textbf{Result on UID~\cite{hu2021unsupervised} comparison with Hu}~\et~\cite{hu2021unsupervised}. The figure includes: input (left), ours (middle), Hu~\et~\cite{hu2021unsupervised} (right).}
\vspace{-1mm}
\label{fig:compare-Hu}
\end{figure}

\begin{figure}[!t]
\centering
\animategraphics[scale=0.37]{5}{gif/iccv2021yu/iccv2021yu_}{45}{56}
\caption{\textbf{Result on H36M comparison with Yu}~\et~\cite{yu2021towards}. The figure includes: input (left), ours (middle), Yu~\et~\cite{yu2021towards} (right). {Red skeleton is prediction, green skeleton is ground truth.}}
\vspace{-1mm}
\label{fig:compare-yu}
\end{figure}


\subsection{Ablation study}

\subsubsection{Ablation on round of co-evolution}
We then analysis how the co-evolving round improves the performance of each component (estimator $\mathcal{P}$, imitator $\mathcal{I}$, hallucinator $\mathcal{H}$).
To demonstrate the improvement, we select three evaluation metrics for each component.
For estimator, we evaluate the trained model $\mathcal{P}$ on H36M test set and report the MPJPE as evaluation metric.
For imitator, we evaluate the trained policy $\mathcal{I}$ on GT 3D reference motion (H36M) to measure the number of termination (\eg, fall down) as evaluation metric.
For hallucinator, we evaluate the trained model $\mathcal{H}$ on GT 3D data (Walking scenario~\cite{harvey2020robust} in H36M) for intermediate pose completion. We measure the MPJPE of pose and root position as evaluation metric.
We involve an {oracle} by training each model using the GT data directly as showed in the last row of Table~\ref{tab:co-evol}.
Through iterative co-evolving, the performance of estimator $\mathcal{P}$, imitator $\mathcal{I}$, hallucinator $\mathcal{H}$ are improved and getting closer to the result which is trained with GT data.
We further provide visualization result for imitator $\mathcal{I}$ (Fig.~\ref{fig:vis-rl}), hallucinator $\mathcal{H}$ (Fig.~\ref{fig:vis-rib}).
This result shows that the imitator $\mathcal{I}$ and hallucinator $\mathcal{H}$ co-evolved by our \nameofmethod{} without using 3D data achieve a comparable performance compared with the {oracle} trained with GT 3D data.

\begin{table}[h]
	\small
	\centering
	\setlength{\tabcolsep}{1mm}
	\begin{tabular}{c|c|c|cc}
		\specialrule{1pt}{1pt}{1pt}
		Round & $\mathcal{P}$ & $\mathcal{I}$ &  \multicolumn{2}{c}{$\mathcal{H}$}  \\
		Num. & P1~(pose) & Termination Num. &  P1~(pose) & P1~(root)  \\
		\hline
		0  & 193.6 & - & - &  -  \\ 
		1  & 112.2 & 928 & - &  -  \\ 
		2  & 77.8 & 280 & 71.4 &  62.6  \\ 
		3  & 68.2 & 132 & 67.3 &  54.0  \\ 
		\hline
		Oracle  & 37.2 & 81 & 53.0 &  33.7  \\ 
		\specialrule{1pt}{1pt}{2pt}	
	\end{tabular}
	\vspace{-1mm}
	\caption{\textbf{Results on co-evolving for estimator $\mathcal{P}$, imitator $\mathcal{I}$, hallucinator $\mathcal{H}$}. Note that round 0 is the \textbf{Loop starting}, and we involve hallucinator $\mathcal{H}$ after round one to ensure the quality of initial pose estimation.} \label{tab:co-evol}
	\vspace{-3mm}
\end{table} 


\begin{figure}[!t]
\centering
\animategraphics[scale=0.37]{5}{gif/3dhp/ts5_}{48}{65}
\caption{\textbf{Result on 3DHP compared with Ground Truth}. The figure includes: input (left), ours (middle), ground truth (right).}
\vspace{-1mm}
\label{fig:vis-3dhp}
\end{figure}

\begin{figure}[!t]
\centering
\animategraphics[scale=0.37]{5}{gif/3dpw/fencing}{28}{44}
\caption{\textbf{Result on 3DPW compared with Ground Truth}. The figure includes: input (left), ours (middle), ground truth (right).}
\vspace{-1mm}
\label{fig:vis-3dpw}
\end{figure}


\begin{figure}[!t]
\centering
\animategraphics[scale=0.37]{5}{gif/video2rl/video2rl_}{10}{30}
\caption{\textbf{Results on co-evolving for imitator $\mathcal{I}$}. The figure includes: video source (left), our co-evolving result (middle), oracle trained with ground truth data(right).}
\vspace{-1mm}
\label{fig:vis-rl}
\end{figure}

\begin{figure}[!t]
\centering
\animategraphics[scale=0.37]{5}{gif/rib/img_008_}{1}{25}
\caption{\textbf{Results on co-evolving for hallucinator $\mathcal{H}$}. The figure includes: ground truth (left), our co-evolving result (middle), oracle trained with ground truth data(right). }
\vspace{-1mm}
\label{fig:vis-rib}
\end{figure}

\subsubsection{Ablation on amount of data usage}
To study how the amount of data affects the performance, we construct an ablation experiment with limited 2D pose data. As shown in Table~\ref{tab:aba-sub}, we gradually involve more data in our method, (\ie, S1, S1+S5, S1+S5+S6+S7+S8). Result shows that the performance of \nameofmethod{} can be improved gradually by adding more 2D pose data in both intra and cross-dataset scenarios.

\begin{table}[h]
	\small
	\centering
	\setlength{\tabcolsep}{1mm}
	\begin{tabular}{l|c|c|c|c}
		\specialrule{1pt}{1pt}{1pt}
		Mode & Sub  & H36M & 3DHP & 3DPW  \\
		\hline
		Self & S1 & 89.2 & 94.0  & 135.8   \\
		Self & S1,S5 & 81.9 & 83.5  & 128.6   \\
		Self & S1,S5,S6,S7,S8 & 68.2 & 79.5  & 115.0   \\
		\specialrule{1pt}{1pt}{2pt}	
	\end{tabular}
	\vspace{-1mm}
	\caption{\textbf{Results on ablation amount of data} in terms of MPJPE.}	\label{tab:aba-sub}
\end{table}


	



\section{Conclusion}
\label{sec:conclusion}
{In this work we present a novel framework \nameofmethod{} for self-supervised 3D pose estimation, which is achieved by a
co-evolution strategy of a pose estimator, imitator, and hallucinator.
These three components, complement and strength one another through a dual-loop strategy as the training procedure. The framework enables generating diverse and plausible motion data, which help train superior pose estimator.
Experiments on varies benchmarks demonstrate that \nameofmethod{} yields encouraging results. It outperforms the state of the art self-supervised approaches and even competes with fully-supervised approaches.}


\noindent \textbf{Limitations} 
The major limitation is that our pipeline suffers low training efficiency, \eg, it takes 7 days to train for 3 rounds on a machine with a Intel Xeon Gold 6278C CPU and a Tesla T4 GPU. The reason is that the imitator ($\mathcal{I}$) is implemented with CPU-based reinforcement learning (RL) and the hallucinator ($\mathcal{H}$) is instantiated with RNN architecture. In the future, we will explore GPU-based RL implementation and more efficient hallucinator architecture (\eg, transformer) to speed up the training process.




\noindent \textbf{Acknowledgement} 
This project is supported by
NUS Faculty Research Committee Grant~(WBS: A-0009440-00-00)
and NUS Advanced Research and Technology Innovation Centre (Project Reference ECT-RP2).
Kehong would like to thank Ye Yuan for the discussion.

{\small
\bibliographystyle{ieee_fullname}
\bibliography{egbib}

\begin{thebibliography}{10}\itemsep=-1pt

\bibitem{barsoum2018hp}
Emad Barsoum, John Kender, and Zicheng Liu.
\newblock Hp-gan: Probabilistic 3d human motion prediction via gan.
\newblock In {\em {CVPRw}}, 2018.

\bibitem{battan2021glocalnet}
Neeraj Battan, Yudhik Agrawal, Sai~Soorya Rao, Aman Goel, and Avinash Sharma.
\newblock Glocalnet: Class-aware long-term human motion synthesis.
\newblock In {\em WACV}, 2021.

\bibitem{butepage2017deep}
Judith Butepage, Michael~J Black, Danica Kragic, and Hedvig Kjellstrom.
\newblock Deep representation learning for human motion prediction and
  classification.
\newblock In {\em {CVPR}}, 2017.

\bibitem{chen2019unsupervised}
Ching-Hang Chen, Ambrish Tyagi, Amit Agrawal, Dylan Drover, Rohith MV, Stefan
  Stojanov, and James~M Rehg.
\newblock Unsupervised 3d pose estimation with geometric self-supervision.
\newblock In {\em {CVPR}}, 2019.

\bibitem{chengpnerf}
Mingfei Chen, Jianfeng Zhang, Xiangyu Xu, Lijuan Liu, Jiashi Feng, and
  Shuicheng Yan.
\newblock Geometry-guided progressive nerf for generalizable and efficient
  neural human rendering.
\newblock {\em arXiv}, 2021.

\bibitem{chen2016synthesizing}
Wenzheng Chen, Huan Wang, Yangyan Li, Hao Su, Zhenhua Wang, Changhe Tu, Dani
  Lischinski, Daniel Cohen-Or, and Baoquan Chen.
\newblock Synthesizing training images for boosting human 3d pose estimation.
\newblock In {\em {3DV}}, 2016.

\bibitem{drover2018can}
Dylan Drover, Rohith MV, Ching-Hang Chen, Amit Agrawal, Ambrish Tyagi, and Cong
  Phuoc~Huynh.
\newblock Can 3d pose be learned from 2d projections alone?
\newblock In {\em {ECCVw}}, 2018.

\bibitem{duan2021ssmc}
Yinglin Duan, Tianyang Shi, Zhengxia Zou, Yenan Lin, Zhehui Qian, Bohan Zhang,
  and Yi Yuan.
\newblock Single-shot motion completion with transformer.
\newblock {\em arXiv}, 2021.

\bibitem{gong2021poseaug}
Kehong Gong, Jianfeng Zhang, and Jiashi Feng.
\newblock Poseaug: A differentiable pose augmentation framework for 3d human
  pose estimation.
\newblock In {\em {CVPR}}, 2021.

\bibitem{gui2018adversarial}
Liang-Yan Gui, Yu-Xiong Wang, Xiaodan Liang, and Jos{\'e}~MF Moura.
\newblock Adversarial geometry-aware human motion prediction.
\newblock In {\em {ECCV}}, 2018.

\bibitem{harvey2018recurrent}
F{\'e}lix~G Harvey and Christopher Pal.
\newblock Recurrent transition networks for character locomotion.
\newblock In {\em {ACM Transactions on Graphics}}, 2018.

\bibitem{harvey2020robust}
F{\'e}lix~G Harvey, Mike Yurick, Derek Nowrouzezahrai, and Christopher Pal.
\newblock Robust motion in-betweening.
\newblock In {\em {ACM Trans. on Graphics}}, 2020.

\bibitem{hernandez2019human}
Alejandro Hernandez, Jurgen Gall, and Francesc Moreno-Noguer.
\newblock Human motion prediction via spatio-temporal inpainting.
\newblock In {\em {ICCV}}, 2019.

\bibitem{holden2016deep}
Daniel Holden, Jun Saito, and Taku Komura.
\newblock A deep learning framework for character motion synthesis and editing.
\newblock In {\em {ACM Trans. on Graphics}}, 2016.

\bibitem{tung_aign}
Tung Hsiao-Yu~Fish, Harley Adam~W, Seto William, and Fragkiadaki Katerina.
\newblock Adversarial inverse graphics networks: {L}earning 2d-to-3d lifting
  and image-to-image translation from unpaired supervision.
\newblock In {\em {ICCV}}, 2017.

\bibitem{hu2021unsupervised}
Xiaodan Hu and Narendra Ahuja.
\newblock Unsupervised 3d pose estimation for hierarchical dance video
  recognition.
\newblock In {\em {ICCV}}, 2021.

\bibitem{ionescu2014human3}
Catalin Ionescu, Dragos Papava, Vlad Olaru, and Cristian Sminchisescu.
\newblock Human3. 6m: Large scale datasets and predictive methods for 3d human
  sensing in natural environments.
\newblock {\em {IEEE Trans. on Pattern Analysis and Machine Intelligence}},
  36(7):1325--1339, 2014.

\bibitem{iqbal2020weakly}
Umar Iqbal, Pavlo Molchanov, and Jan Kautz.
\newblock Weakly-supervised 3d human pose learning via multi-view images in the
  wild.
\newblock In {\em {CVPR}}, 2020.

\bibitem{jiang2020coherent}
Wen Jiang, Nikos Kolotouros, Georgios Pavlakos, Xiaowei Zhou, and Kostas
  Daniilidis.
\newblock Coherent reconstruction of multiple humans from a single image.
\newblock In {\em {CVPR}}, 2020.

\bibitem{joo2020exemplar}
Hanbyul Joo, Natalia Neverova, and Andrea Vedaldi.
\newblock Exemplar fine-tuning for 3d human pose fitting towards in-the-wild 3d
  human pose estimation.
\newblock {\em {3DV}}, 2020.

\bibitem{kaufmann2020convolutional}
Manuel Kaufmann, Emre Aksan, Jie Song, Fabrizio Pece, Remo Ziegler, and Otmar
  Hilliges.
\newblock Convolutional autoencoders for human motion infilling.
\newblock {\em {3DV}}, 2020.

\bibitem{kocabas2020vibe}
Muhammed Kocabas, Nikos Athanasiou, and Michael~J Black.
\newblock Vibe: Video inference for human body pose and shape estimation.
\newblock In {\em {CVPR}}, 2020.

\bibitem{kocabas2019epipolar}
Muhammed Kocabas, Salih Karagoz, and Emre Akbas.
\newblock Self-supervised learning of 3d human pose using multi-view geometry.
\newblock In {\em {CVPR}}, 2019.

\bibitem{kolotouros2019spin}
Nikos Kolotouros, Georgios Pavlakos, Michael~J Black, and Kostas Daniilidis.
\newblock Learning to reconstruct 3d human pose and shape via model-fitting in
  the loop.
\newblock In {\em {ICCV}}, 2019.

\bibitem{kovar2008motion}
Lucas Kovar, Michael Gleicher, and Fr{\'e}d{\'e}ric Pighin.
\newblock Motion graphs.
\newblock In {\em {ACM Transactions on Graphics}}, 2008.

\bibitem{kundu2020self}
Jogendra~Nath Kundu, Siddharth Seth, Varun Jampani, Mugalodi Rakesh,
  R~Venkatesh Babu, and Anirban Chakraborty.
\newblock Self-supervised 3d human pose estimation via part guided novel image
  synthesis.
\newblock In {\em {CVPR}}, 2020.

\bibitem{kundu2020kinematic}
Jogendra~Nath Kundu, Siddharth Seth, MV Rahul, Mugalodi Rakesh, Venkatesh~Babu
  Radhakrishnan, and Anirban Chakraborty.
\newblock Kinematic-structure-preserved representation for unsupervised 3d
  human pose estimation.
\newblock In {\em {AAAI}}, 2020.

\bibitem{li2018convolutional}
Chen Li, Zhen Zhang, Wee~Sun Lee, and Gim~Hee Lee.
\newblock Convolutional sequence to sequence model for human dynamics.
\newblock In {\em {CVPR}}, 2018.

\bibitem{Li_2020_CVPR}
Shichao Li, Lei Ke, Kevin Pratama, Yu-Wing Tai, Chi-Keung Tang, and Kwang-Ting
  Cheng.
\newblock Cascaded deep monocular 3d human pose estimation with evolutionary
  training data.
\newblock In {\em {CVPR}}, 2020.

\bibitem{li2019boosting}
Zhi Li, Xuan Wang, Fei Wang, and Peilin Jiang.
\newblock On boosting single-frame 3d human pose estimation via monocular
  videos.
\newblock In {\em {ICCV}}, 2019.

\bibitem{liu2021spatt}
Ting Liu, Jianfeng Zhang, Xuecheng Nie, Yunchao Wei, Shikui Wei, Yao Zhao, and
  Jiashi Feng.
\newblock Spatial-aware texture transformer for high-fidelity garment transfer.
\newblock In {\em {IEEE Trans. on Image Processing}}, 2021.

\bibitem{martinez2017human}
Julieta Martinez, Michael~J Black, and Javier Romero.
\newblock On human motion prediction using recurrent neural networks.
\newblock In {\em {CVPR}}, 2017.

\bibitem{martinez2017simple}
Julieta Martinez, Rayat Hossain, Javier Romero, and James~J Little.
\newblock A simple yet effective baseline for 3d human pose estimation.
\newblock In {\em {ICCV}}, 2017.

\bibitem{mehta2017vnect}
Dushyant Mehta, Srinath Sridhar, Oleksandr Sotnychenko, Helge Rhodin, Mohammad
  Shafiei, Hans-Peter Seidel, Weipeng Xu, Dan Casas, and Christian Theobalt.
\newblock Vnect: Real-time 3d human pose estimation with a single rgb camera.
\newblock {\em {ACM Trans. on Graphics}}, 36(4):44, 2017.

\bibitem{buttner2015motion}
Buttner Michael.
\newblock Motion matching-the road to next-gen animation.
\newblock In {\em Nucl. ai}, 2015.

\bibitem{mitra2020multiview}
Rahul Mitra, Nitesh~B Gundavarapu, Abhishek Sharma, and Arjun Jain.
\newblock Multiview-consistent semi-supervised learning for 3d human pose
  estimation.
\newblock In {\em {CVPR}}, 2020.

\bibitem{nie2019spm}
Xuecheng Nie, Jianfeng Zhang, Shuicheng Yan, and Jiashi Feng.
\newblock Single-stage multi-person pose machines.
\newblock In {\em {ICCV}}, 2019.

\bibitem{pavllo2019videopose3d}
Dario Pavllo, Christoph Feichtenhofer, David Grangier, and Michael Auli.
\newblock 3d human pose estimation in video with temporal convolutions and
  semi-supervised training.
\newblock In {\em {CVPR}}, 2019.

\bibitem{pavllo2018quaternet}
Dario Pavllo, David Grangier, and Michael Auli.
\newblock Quaternet: A quaternion-based recurrent model for human motion.
\newblock {\em arXiv}, 2018.

\bibitem{peng2018deepmimic}
Xue~Bin Peng, Pieter Abbeel, Sergey Levine, and Michiel van~de Panne.
\newblock Deepmimic: Example-guided deep reinforcement learning of
  physics-based character skills.
\newblock In {\em {ACM Trans. on Graphics}}, 2018.

\bibitem{peng2018sfv}
Xue~Bin Peng, Angjoo Kanazawa, Jitendra Malik, Pieter Abbeel, and Sergey
  Levine.
\newblock Sfv: Reinforcement learning of physical skills from videos.
\newblock In {\em {ACM Trans. on Graphics}}, 2018.

\bibitem{peng2017learning}
Xue~Bin Peng and Michiel van~de Panne.
\newblock Learning locomotion skills using deeprl: Does the choice of action
  space matter?
\newblock In {\em ACM}, 2017.

\bibitem{rempe2020contact}
Davis Rempe, Leonidas~J Guibas, Aaron Hertzmann, Bryan Russell, Ruben Villegas,
  and Jimei Yang.
\newblock Contact and human dynamics from monocular video.
\newblock In {\em {ECCV}}, 2020.

\bibitem{rhodin2018unsupervised}
Helge Rhodin, Mathieu Salzmann, and Pascal Fua.
\newblock Unsupervised geometry-aware representation for 3d human pose
  estimation.
\newblock In {\em {ECCV}}, 2018.

\bibitem{shimada2021neural}
Soshi Shimada, Vladislav Golyanik, Weipeng Xu, Patrick P{\'e}rez, and Christian
  Theobalt.
\newblock Neural monocular 3d human motion capture with physical awareness.
\newblock 2021.

\bibitem{shimada2020physcap}
Soshi Shimada, Vladislav Golyanik, Weipeng Xu, and Christian Theobalt.
\newblock Physcap: Physically plausible monocular 3d motion capture in real
  time.
\newblock In {\em {ACM Trans. on Graphics}}, 2020.

\bibitem{si2019attention}
Chenyang Si, Wentao Chen, Wei Wang, Liang Wang, and Tieniu Tan.
\newblock An attention enhanced graph convolutional lstm network for
  skeleton-based action recognition.
\newblock In {\em {CVPR}}, 2019.

\bibitem{sun2018integral}
Xiao Sun, Bin Xiao, Fangyin Wei, Shuang Liang, and Yichen Wei.
\newblock Integral human pose regression.
\newblock In {\em {ECCV}}, 2018.

\bibitem{sun2019human}
Yu Sun, Yun Ye, Wu Liu, Wenpeng Gao, Yili Fu, and Tao Mei.
\newblock Human mesh recovery from monocular images via a skeleton-disentangled
  representation.
\newblock In {\em {ICCV}}, 2019.

\bibitem{vonMarcard2018}
Timo von Marcard, Roberto Henschel, Michael Black, Bodo Rosenhahn, and Gerard
  Pons-Moll.
\newblock Recovering accurate 3d human pose in the wild using imus and a moving
  camera.
\newblock In {\em {ECCV}}, 2018.

\bibitem{wandt2021canonpose}
Bastian Wandt, Marco Rudolph, Petrissa Zell, Helge Rhodin, and Bodo Rosenhahn.
\newblock Canonpose: Self-supervised monocular 3d human pose estimation in the
  wild.
\newblock In {\em {CVPR}}, 2021.

\bibitem{WangJueICCV19}
Jue Wang, Shaoli Huang, Xinchao Wang, and Dacheng Tao.
\newblock Not all parts are created equal: 3d pose estimation by modelling
  bi-directional dependencies of body parts.
\newblock In {\em {ICCV}}, 2019.

\bibitem{wang2021mvp}
Tao Wang, Jianfeng Zhang, Yujun Cai, Shuicheng Yan, and Jiashi Feng.
\newblock Direct multi-view multi-person 3d pose estimation.
\newblock In {\em {NeurIPS}}, 2021.

\bibitem{wang2020predicting}
Zhe Wang, Daeyun Shin, and Charless~C Fowlkes.
\newblock Predicting camera viewpoint improves cross-dataset generalization for
  3d human pose estimation.
\newblock In {\em {ECCVw}}, 2020.

\bibitem{wang2020learning}
Zhenyi Wang, Ping Yu, Yang Zhao, Ruiyi Zhang, Yufan Zhou, Junsong Yuan, and
  Changyou Chen.
\newblock Learning diverse stochastic human-action generators by learning
  smooth latent transitions.
\newblock In {\em {AAAI}}, 2020.

\bibitem{wu2016single}
Jiajun Wu, Tianfan Xue, Joseph~J Lim, Yuandong Tian, Joshua~B Tenenbaum,
  Antonio Torralba, and William~T Freeman.
\newblock Single image 3d interpreter network.
\newblock In {\em {ECCV}}, 2016.

\bibitem{xie2021physics}
Kevin Xie, Tingwu Wang, Umar Iqbal, Yunrong Guo, Sanja Fidler, and Florian
  Shkurti.
\newblock Physics-based human motion estimation and synthesis from videos.
\newblock In {\em {CVPR}}, 2021.

\bibitem{yan2018spatial}
Sijie Yan, Yuanjun Xiong, and Dahua Lin.
\newblock Spatial temporal graph convolutional networks for skeleton-based
  action recognition.
\newblock In {\em {AAAI}}, 2018.

\bibitem{Yang2020Distill}
Yiding Yang, Jiayan Qiu, Mingli Song, Dacheng Tao, and Xinchao Wang.
\newblock Distilling knowledge from graph convolutional networks.
\newblock In {\em {CVPR}}, 2020.

\bibitem{Yang2021Pose}
Yiding Yang, Zhou Ren, Haoxiang Li, Chunluan Zhou, Xinchao Wang, and Gang Hua.
\newblock Learning dynamics via graph neural networks for human pose estimation
  and tracking.
\newblock In {\em {CVPR}}, 2021.

\bibitem{yu2021towards}
Zhenbo Yu, Bingbing Ni, Jingwei Xu, Junjie Wang, Chenglong Zhao, and Wenjun
  Zhang.
\newblock Towards alleviating the modeling ambiguity of unsupervised monocular
  3d human pose estimation.
\newblock In {\em {ICCV}}, 2021.

\bibitem{yuan2020dlow}
Ye Yuan and Kris Kitani.
\newblock Dlow: Diversifying latent flows for diverse human motion prediction.
\newblock In {\em {ECCV}}, 2020.

\bibitem{yuan2020residual}
Ye Yuan and Kris Kitani.
\newblock Residual force control for agile human behavior imitation and
  extended motion synthesis.
\newblock In {\em {NeurIPS}}, 2020.

\bibitem{yuan2021simpoe}
Ye Yuan, Shih-En Wei, Tomas Simon, Kris Kitani, and Jason Saragih.
\newblock Simpoe: Simulated character control for 3d human pose estimation.
\newblock In {\em {CVPR}}, 2021.

\bibitem{zhang2020inference}
Jianfeng Zhang, Xuecheng Nie, and Jiashi Feng.
\newblock Inference stage optimization for cross-scenario 3d human pose
  estimation.
\newblock In {\em {NeurIPS}}, 2020.

\bibitem{zhang2021bmp}
Jianfeng Zhang, Dongdong Yu, Jun~Hao Liew, Xuecheng Nie, and Jiashi Feng.
\newblock Body meshes as points.
\newblock In {\em {CVPR}}, 2021.

\bibitem{zhang2020we}
Yan Zhang, Michael~J Black, and Siyu Tang.
\newblock We are more than our joints: Predicting how 3d bodies move.
\newblock {\em {CVPR}}, 2021.

\bibitem{zhou2017towards}
Xingyi Zhou, Qixing Huang, Xiao Sun, Xiangyang Xue, and Yichen Wei.
\newblock Towards 3d human pose estimation in the wild: a weakly-supervised
  approach.
\newblock In {\em {ICCV}}, 2017.

\end{thebibliography}
}


\end{document}